# Natural Language Processing for Aligning Readings with Learning Objectives and Assessing Topic Balance in Dental Education

Farzana Yasmin, Tahereh Firoozi, Rojin Adabdokht

**Abstract.** Curriculum teams often lack time to check whether assigned readings support learning objectives (LOs) and whether topic coverage is balanced. We present a natural language processing (NLP) workflow to support evidence-informed curation of reading lists. The workflow combines three components. First, PubMedBERT sentence embeddings and cosine similarity retrieve and rank texts that are semantically close to an LO. Second, NLI provides an evidence check to filter weak or non-supportive matches. Third, topic modeling summarizes coverage in the retrieved corpus to flag gaps and overemphasis. We tested the approach in dental education by defining 13 course-level LOs (caries, periodontitis, dental implants, endodontics) and using a PubMed-derived proxy reading corpus that yielded 3,837 LO–abstract pairs. Across the 13 LOs, median cosine similarity ranged from 0.86 to 0.91, indicating generally high semantic similarity between the LOs and retrieved abstracts. Latent Dirichlet Allocation suggested broad but uneven coverage, with one or two dominant topics per domain. This can impact constructive alignment because under-coverage of topics can create learning gaps relative to the intended LOs and assessments. Similarity alone was insufficient for selecting readings. Among high-similarity pairs (cosine 0.90-1.00), a Natural Language Inference (NLI) evidence check classified 22.9% as neutral or contradictory, indicating that cosine retrieval can overestimate LO support. Cosine similarity and entailment probability were moderately correlated (Spearman $\rho = 0.43$; Pearson $r = 0.47$), which supports NLI as a post-retrieval verification step. In a 49-pair expert-labeled validation set, cosine-based retrieval (cosine $\geq 0.90$) achieved 0.69 accuracy, while the combined cosine and NLI approach achieved 0.73 accuracy against expert alignment labels. Overall, the workflow shows how AI can support curriculum alignment decisions by reducing false positives and improving precision when selecting readings that match intended LOs.

# 1 Introduction

When educators say a curriculum should be aligned, they mean that learning objectives (LOs), teaching activities, readings, and assessments should all point to the same intended outcomes [2, 3]. In practice, parts of a course often change at different speeds. Reading lists and assessments can be updated annually, whereas formal objectives and program documents change more slowly. When this happens, students may be assessed on material that is not well supported by assigned readings, and coverage can become uneven across domains. Curriculum mapping can detect these gaps and redundancies, but doing it well usually requires coordinated review teams and sustained effort [2-4].

This study tests whether natural language processing (NLP) can support faster, more scalable curriculum review in dental education. We focus on four core oral health domains commonly represented in predoctoral curricula (caries, periodontitis, dental implants, and endodontics) and define 13 course-level LOs across these domains (caries: n = 4; periodontitis: n = 3; dental implants: n = 3; endodontics: n = 3). We treat PubMed abstracts as a proxy pool of assignable research readings. Then, we quantify (a) which texts most strongly support each LO and (b) how evenly the reading pool covers the intended domains. We use biomedical BERT-based sentence embeddings for similarity matching [23, 24]. We also add an evidence-support check using natural language inference (NLI), which aims to reduce on-topic but not actually supportive matches [14-17]. Finally, we use topic modeling to summarize coverage patterns that may signal imbalance [20, 25]. To assess performance, a dentistry expert manually labeled 49 LO-abstract pairs, which we use as a small validation set to estimate agreement and error rates for similarity-only retrieval and the NLI evidence check.

**Research questions.**

**Alignment**. Can NLP-based similarity-only matching identify alignment between the reading list and LO, and does an NLI-based evidence check reduce false positives compared with similarity-only matching?

**Coverage**. Can topic modeling summarize topic coverage across the LO-linked reading corpus in a way that is interpretable and useful as descriptive evidence for curriculum review?

# 2 Literature Review

In the classic objective-based view, Ralph W. Tyler framed curriculum design around four practical questions: what outcomes we want, what learning experiences could produce them, how to organize those experiences, and how to evaluate whether the outcomes were reached [1]. Building on this foundation, John Biggs described constructive alignment as designing LOs, teaching activities, and assessments together so they support the same intended outcomes [2]. In medical and health-professions education, Ronald M. Harden (2001) argued that curriculum mapping makes this alignment more transparent by showing what, when, and how content is taught, and how it is assessed [3]. Together, these perspectives motivate a simple problem: alignment is a goal but verifying it across many LOs and resources is labor-intensive [3,4].

In dental education, curriculum mapping is commonly used to link courses and assessments to competency frameworks, identify gaps, and support review cycles [5-10]. Examples include electronic competency maps and course-level mapping studies in dental programs. These studies are valuable, but they typically rely on manual review of course documents and assessment materials, often by faculty or analysts. As a result, they usually do not quantify whether specific readings provide strong support for specific LOs, especially when the same idea can be expressed using different clinical or research terms.

A newer body of work uses NLP to automate parts of alignment. Zaki et al. (2023) built an NLP-based system that maps course learning outcomes to program LOs using modern language features, including BERT-based representations, and they reported high agreement with expert mappings in two higher education programs [11]. Butterfuss and Doran (2025) applied language-model embeddings to align content standards with assessment materials and showed that embeddings can help humans identify poorly aligned items [12]. Kilinc et al. (2025) fine-tuned transformer models to classify whether course-level intended LOs align with broader graduate LOs and used this to support constructive alignment reviews at scale [13]. Together, these studies show that embedding-based NLP methods can support outcome-to-outcome and standard-to-item alignment in general education settings, but they have not been applied to dental curricula or to reading lists.

Recent work in educational measurement shows how NLP can reduce manual effort in alignment tasks. Embedding-based approaches have been used to map course outcomes to program outcomes and to support content mapping between sets of standards [11-13]. These approaches are good at finding text that is topically similar. Our added NLI step serves a different purpose: determining whether a text supports a claim, not just whether it is about the same topic. In NLP, this support vs not support idea is central to NLI. It grew out of recognising textual entailment work (for example, the Recognizing Textual Entailment challenges) and later benchmark datasets like SNLI and MultiNLI that label pairs as entailment, contradiction, or neutral [14-16]. It also appears in domain-specific resources such as MedNLI and scientific evidence tasks such as SciFact, and in clinical-trial evidence settings such as NLI4CT [17-19].

Unsupervised topic modeling has also been used to study curriculum content and topic balance. Takizawa (2023) used LDA-based topic modeling on pre-clerkship curricular documents to map content to competencies and quantify coverage, including tracking an emerging area of interest (gender identity) over time [20]. Other studies use topic models on educational texts, such as discussion forum posts, to show how themes in learning shift over time and to detect patterns that are hard to see through manual review, as in Barker et al. (2024) on motivations for enrolling in online professional development [21]. However, these topic modeling approaches are usually applied to institutional documents or educational discourse alone and are not directly combined with explicit LOs.

Taken together, the gap is clear. Dental curriculum mapping has strong frameworks and practical use, but it rarely tests whether the research literature that might be assigned as readings provides direct, LO-specific support at scale. NLP alignment work has shown promise in other education settings, and NLI offers a principled way to reduce false positives by checking evidence support. Topic modeling offers a complementary lens by summarizing coverage and redundancy. This study brings these strands together to produce scalable evidence for curriculum review using LOs and a large set of research abstracts as a proxy reading pool.

# 3 Methods

Our first research question is whether NLP-based similarity matching can identify readings that clearly support each LO, and whether an NLI evidence check reduces false positives. We use PubMedBERT for similarity-only matching and a biomedical NLI model for evidence checking. We focus on 13 LOs across four dental domains, build a PubMed abstract–LO dataset, and evaluate performance using an expert-labeled validation set. Our second research question asks whether topic coverage is balanced across domains. We use LDA for topic modeling to summarize themes in the corpus and compare topic distributions across domains.

## 3.1 Data Description

To test whether readings can support each LO, we constructed a proxy reading corpus of PubMed abstracts in place of a course reading list. We focused on four oral health domains aligned with our LOs: caries, periodontitis, dental implants, and endodontics. We ran 13 LO-specific PubMed queries, each combining a domain anchor (MeSH major topic when available plus title and abstract keywords) with LO-specific concept terms, then restricted results to English-language records with abstracts published between 2015 and 2025.

Domain-specific concept blocks emphasized (i) diet and biofilm mechanisms and caries management decisions in caries, (ii) pathogenesis, risk factors, clinical and radiographic signs, and maintenance planning in periodontitis, (iii) implant site assessment, placement and restoration planning, and complication follow-up in implants, and (iv) pulp pain and diagnostic testing plus cleaning and obturation decisions and outcomes in endodontics.

We excluded editorials, letters, comments, errata, retractions, and non-human studies. For each query, we exported PMID, title, abstract, journal name, and publication year into a separate file and then combined the LO-specific files into a single dataset with domain and LO identifiers. We removed exact duplicate records, merged the data with a unique PMID-to-abstract lookup, and removed rows with missing or empty abstracts. We retained the same PMID when it appeared under different LOs because the unit of analysis was the LO–abstract pair. For caries, the retrieval counts were LO1 = 563, LO2 = 388, LO3 = 110, and LO4 = 504. For periodontitis, they were LO1 = 604, LO2 = 108, and LO3 = 40. For dental implants, they were LO1 = 109, LO2 = 393, and LO3 = 572. For endodontics, they were LO1 = 160, LO2 = 135, and LO3 = 768. After these cleaning and merging steps, the final analytic dataset contained 3,837 LO–abstract pairs.

In parallel, we defined 13 LOs spanning four oral health domains (caries, periodontitis, dental implants, endodontics). The objectives were written at a predoctoral level and structured to reflect common clinical competencies in each domain, including recognizing signs and symptoms, selecting and interpreting basic tests, assessing risk, choosing preventive versus operative management, planning procedures and follow-up, and identifying indications for referral. Each LO was intentionally concise and framed as an observable skill using action verbs. We stored the objectives in a structured table with domain and LO identifiers and the full LO text, enabling one-to-one mapping between each objective and the PubMed retrieval and downstream LO-abstract alignment analyses.

## 4 Data Analysis

We analyzed the PubMed abstract pool (used as a proxy reading list) to quantify LO-reading alignment and topic coverage using PubMedBERT embeddings, an NLI verification step, and LDA topic modeling. Below, we describe each method used in our analysis.

### 4.1 BERT Embedding and Cosine Similarity

For PubMedBERT, we used the sentence-transformers model pritamdeka/S-PubMedBert-MS-MARCO, which is built on microsoft/BiomedNLP-PubMedBERT-base-uncased-abstract-fulltext (also released under the updated BiomedBERT naming) and is fine-tuned on the MS MARCO dataset for information retrieval using the sentence-transformers framework [22, 24]. In this study, we used the model only for inference, meaning we did not further train or fine-tune it on our data. Each input text was passed through the fixed pretrained encoder, and sentence embeddings were obtained using the model's mean-pooling setup to produce a 768-dimensional dense vector.

Using the BERT-based sentence embedding model, we encoded both the LOs and the abstracts into numeric vectors in the same embedding space. For similarity scoring, we concatenated each record's title and abstract before encoding. This provides both a high-level summary signal (title) and the detailed content (abstract), which are useful for representing the overall meaning of a candidate reading. We then computed cosine similarity between each LO vector and every abstract vector. For each LO, we ranked all abstracts by cosine similarity. We defined a similarity threshold of 0.90 to mark an abstract as aligned with that LO. We conducted all analyses in Jupyter Notebook using Python.

### 4.2 Natural Language Inference Evidence Checks

To go beyond topical similarity, we added an NLI step to estimate whether an abstract supports an LO, not just whether it is about the same topic. NLI is a three-class task that predicts whether a premise entails, contradicts, or is neutral with respect to a hypothesis [15].
We used the biomedical NLI model pritamdeka/PubMedBERT-MNLI-MedNLI and ran it in inference-only mode [17]. The model predicts probabilities for the labels entailment, neutral, and contradiction [22].
**Inputs**. For each LO–abstract pair, we defined the premise as the abstract text and the hypothesis as the LO text. Specifically, we reconstructed the premise by merging the cosine-similarity pair file with the source PubMed file using PMID, then concatenating the title and abstract into one string. The LO text was used as the hypothesis after basic whitespace cleaning.
**Computation**. We tokenized premise-hypothesis pairs with the Hugging Face tokenizer and truncated inputs to a maximum of 512 tokens.
**Outputs and scoring**. For each pair, the model returned logits, which we converted to probabilities using softmax. We stored the entailment, neutral, and contradiction probabilities, along with the predicted label. We treated the entailment probability as an "evidence support" score. In downstream analyses, we used this score to filter or re-rank similarity-based matches, aiming to reduce false positives that are on-topic but not actually supportive.

### 4.3 LDA Analysis

To study topic coverage and topic balance, we applied LDA to the 3,837 LO–abstract records in the analytic dataset. For topic modeling, we used abstracts only. LDA is a bag-of-words model, and frequent generic terms can dominate topics if not removed or downweighted [20, 25]. Including titles can increase the influence of broad, high-level terms and can lead to less specific topics, so we kept the LDA input focused on abstract text.

Before fitting the model, we converted the abstracts to lowercase, removed common stop words, and created a document term matrix. We chose the number of topics using a mix of domain knowledge and model diagnostics and then examined the top words and example abstracts for each topic. We used document-level topic probabilities to estimate how often each topic appeared in each domain and to see whether some domains dominated the corpus.

### 4.4 Expert Validation and Performance Evaluation

A domain expert reviewed 50 LO–abstract pairs. One pair had no usable expert rating, so we excluded it, leaving 49 pairs for evaluation. We considered an abstract aligned when it provided meaningful coverage of the main instructional focus of the LO. Partial coverage was acceptable when the expert judged the abstract to be meaningfully relevant to the objective. However, we coded a pair as not aligned when the expert indicated that a central component of the LO was not addressed.
We used a cosine similarity threshold of 0.90 for the BERT-only analysis. For the NLI-only analysis, we used an entailment probability threshold of 0.50 as the binary decision threshold. For the combined analysis, we fixed cosine similarity at 0.90 and evaluated a range of NLI entailment thresholds. The highest F1 scores occurred between 0.30 and 0.40 in this pilot sample, so we used 0.30 for the combined decision rule. We evaluated each method using precision, recall, F1 score, and accuracy (Fig. 1).

# 5 Results

In this section, we report the results of the PubMedBERT retrieval, the NLI verification step, and the LDA topic coverage analysis.

## 5.1 PubMed-based BERT Result

From the PubMed-based BERT analysis, across all four domains, cosine similarity scores for the top matching abstracts were consistently high. For every LO, the median similarity for its best-matched abstracts was roughly between 0.86 and 0.91, and the 90th percentile scores typically ranged from about 0.88 to 0.93 (Table 1). Even the lowest minima among the top matches stayed at or above 0.83. Caries LOs generally showed some of the highest median and upper-tail similarity scores, while several periodontitis and endodontics LOs had lower values. However, all domains still included abstracts with high semantic similarity to their LOs. Overall, this pattern suggests that the public dental literature contains strong candidate readings with meaningful semantic alignment to the defined LOs.

## 5.2 NLI Results

NLI labels shift strongly with cosine similarity. In the 0.8-0.9 range, entailment is only 47.9%, with 41.2% neutral and 10.9% contradiction (Table 2). Even above 0.9, 22.9% of pairs are not entailment (16.8% neutral, 6.1% contradiction), showing that high similarity can still produce on-topic but non-supportive matches.

Cosine similarities from PubMedBERT and NLI entailment probabilities were moderately correlated (Spearman $\rho = 0.429$; Pearson $r = 0.471$; both $p < 0.001$), indicating related but non-redundant signals. This supports using NLI as an evidence check to reduce false positives among high-similarity matches. We merged expert labels with cosine and NLI outputs using (LO index, PMID). One pair was excluded due to a missing/unclear expert label, leaving $n = 49$ for evaluation.

## 5.3 LDA Result

Across the corpus, LDA topic mass is distributed rather than dominated by a single topic. Topic 1 has the highest average proportion (16.6%), followed by Topic 3 (13.5%) and Topic 4 (12.4%). The remaining topics each contribute between 5.2% and 12.1%. Together, the top three topics account for 42.5% of the total mass, suggesting several recurring themes rather than one dominant cluster.

At the domain level, each area draws most of its topic mass from one or two topics, but the patterns differ by domain. Caries is split across Topic 3 (29%) and Topic 4 (28%), with Topic 0 (19%) also contributing meaningfully (Table 3). Dental implants is strongly concentrated in Topic 1 (56%), with Topic 7 (22%) as the second largest topic. Endodontics is dominated by Topic 6 (42%), with Topic 2 (20%) and Topic 8 (20%) adding substantial mass. Periodontitis is mainly represented by Topic 9 (46%) and Topic 5 (31%). Overall, the LDA results suggest that each domain is concentrated in one or two prominent topics, although domains still share topics, which fits the idea of overlapping content rather than isolated silos.

### 5.4 Performance Evaluation

Results indicate that cosine similarity is effective for retrieving candidate readings, while NLI is most useful as a post-retrieval evidence check that reduces false positives. Using a cosine threshold of 0.90, the BERT-only method achieved precision = 0.61, recall = 0.86, F1 = 0.72, and accuracy = 0.69 (TP = 19, FP = 12, FN = 3, TN = 15), indicating high sensitivity to expert-aligned matches but a tendency to include false positives (Table 4). NLI alone at an entailment threshold of 0.50 achieved precision = 0.57, recall = 0.77, F1 = 0.65, and accuracy = 0.63 (TP = 17, FP = 13, FN = 5, TN = 14). The combined BERT + NLI approach (cosine ≥ 0.90 and entailment ≥ 0.30) achieved precision = 0.68, recall = 0.77, F1 = 0.72, and accuracy = 0.73 (TP = 17, FP = 8, FN = 5, TN = 19). Compared with BERT-only retrieval, the combined approach reduced false positives from 12 to 8 and increased precision from 0.61 to 0.68 and accuracy from 0.69 to 0.73, although recall decreased from 0.86 to 0.77. These results suggest that NLI can serve as a post-retrieval evidence check for filtering candidate readings identified through semantic similarity.

Fig. 1 shows how the combined method performs when cosine similarity is fixed at 0.90 and the NLI entailment threshold is varied. As the entailment threshold increases, precision generally increases while recall decreases, reflecting a stricter evidence requirement. Performance is relatively stable between 0.30 and 0.40, where the combined rule achieves its highest F1 scores on the expert-labeled set. At higher thresholds, the method becomes more conservative, with slightly higher precision but lower recall. Overall, the figure shows the trade-off between reducing false positives and retaining aligned readings and supports using NLI as a post-retrieval evidence check.

## 6 Limitations

This study uses PubMed abstracts as a proxy reading pool, rather than a real dental curriculum with course syllabi and assigned readings. Therefore, the alignment and topic coverage estimates reflect what is available in the public dental research literature, not what students read in a specific program. Topic model results also depend on modelling choices, such as preprocessing and the number of topics, and topic labels require human judgment, so some subjectivity remains. In addition, the LOs were written by a single author, and the expert-labeled validation set used to evaluate thresholds is small, so the reported performance and chosen cut-offs should be treated as preliminary. As part of this pilot phase, the domain expert noted that the LOs may need tighter oral-health-specific phrasing and recommended building a more reliable reference set by having two or more domain experts label the same pairs independently. Finally, NLI entailment probabilities may not be well calibrated across domains or LO wording, so NLI thresholds may not transfer without re-tuning on new data.

## 7 Discussion

A key strength of this study is the use of a post-retrieval evidence check to control false positives. Similarity-based retrieval alone can return abstracts that are topically related but do not meaningfully align with the LO. In the expert validation set, adding an NLI evidence check after cosine-based retrieval reduced false positives and improved overall agreement with the expert alignment labels, although recall decreased. This supports the idea that reading-list retrieval should be treated as a two-stage problem: first, retrieve plausible candidates based on semantic similarity, then filter for evidence-based support.

Even without using real course syllabi, the results suggest that recent dental research literature can provide strong textual support for core LOs. PubMedBERT embeddings produced consistently high top-match similarity scores, indicating that for common domains such as caries, dental implants, endodontics, and periodontitis, the public literature contains many candidate readings that align well with typical curriculum expectations. The LDA topic analysis adds a second view on coverage. Each domain drew on one or two dominant topics, and no single topic dominated the entire corpus, which suggests a broad but uneven mix of themes. Together, these findings position the method as a screening and decision-support tool that can narrow large reading pools to a manageable shortlist for expert review, rather than replacing expert judgment.

## 8 Conclusion

This study shows that modern NLP methods can quantify the extent to which a large pool of dental research abstracts aligns with clearly stated LOs and the distribution of topic coverage across domains. Using PubMedBERT embeddings, we ranked candidate readings for each LO, and using LDA, we summarized topic concentration within and across domains. Adding an NLI step provided an evidence-support check that reduced false positives from similarity-only retrieval, which is important when the goal is to recommend readings that meaningfully align with an objective rather than merely share vocabulary. Future work should apply the same pipeline to real course syllabi and assigned readings, expand expert validation, and test how robust the chosen thresholds are across different domains and LO phrasing. With these extensions, the approach can support more evidence-informed curriculum review by making alignment checks faster, more transparent, and easier to repeat over time.

## Acknowledgments
The authors thank Dr. Mark Gierl for his helpful feedback on the interpretation of the study findings.

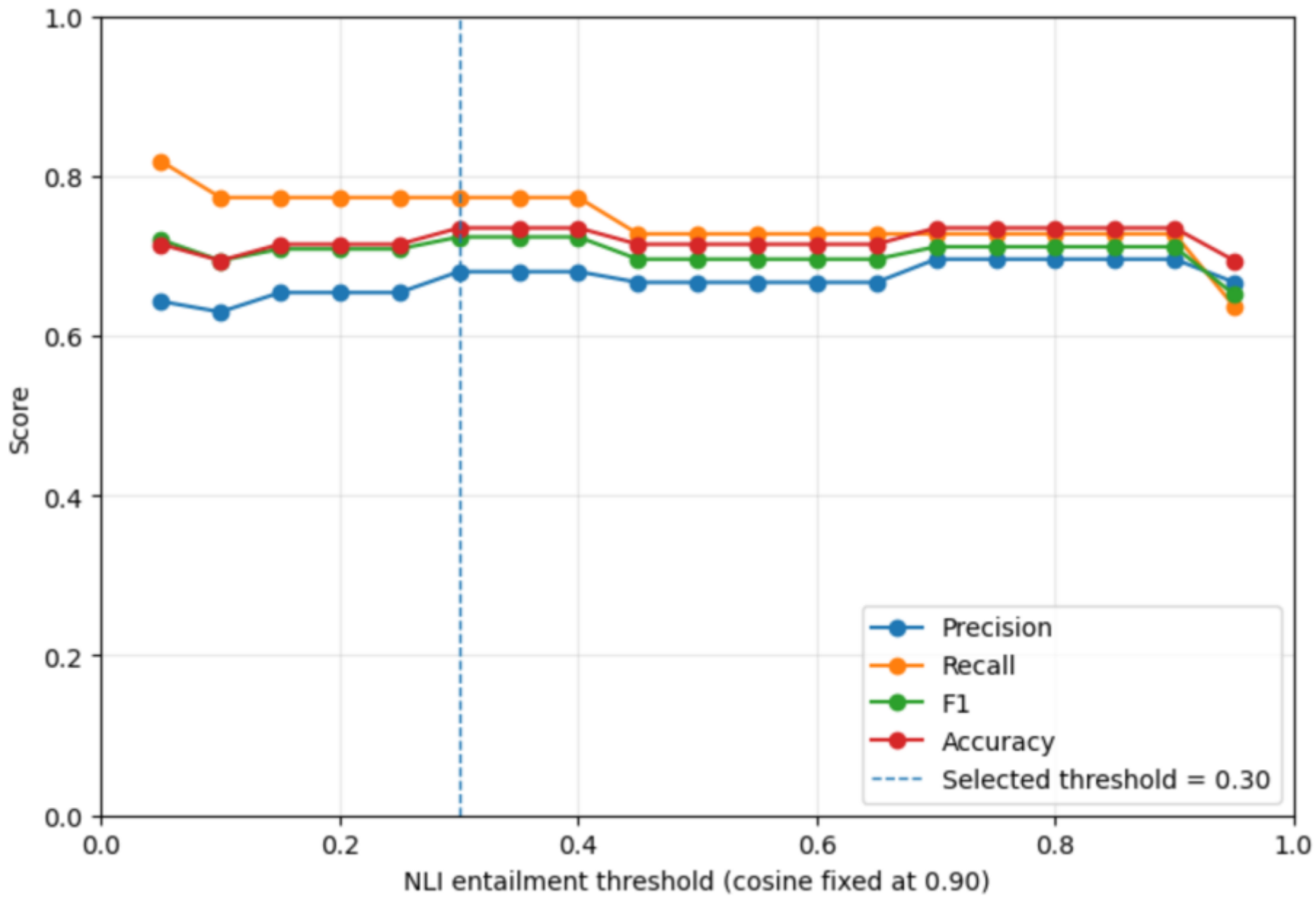

Fig. 1. PubMedBERT + NLI performance trade-off on the expert-labeled set.

**Table 1**. BERT-based cosine similarity for each learning objective in four domains.

| LO ID | Domain | Learning Objectives (text) | Min | Max | Mean | q50 Median | q90 |
|---|---|---|---|---|---|---|---|
| LO1 | Caries | Explain caries biology and the role of diet in disease progression. | 0.85 | 0.96 | 0.90 | 0.90 | 0.92 |
| LO2 | | Detect and diagnose dental caries at patient and tooth levels and assess caries risk. | 0.85 | 0.95 | 0.91 | 0.91 | 0.93 |
| LO3 | | Emphasize prevention and select non-invasive versus invasive management. | 0.84 | 0.92 | 0.88 | 0.88 | 0.89 |
| LO4 | | Identify indications and techniques for removal of carious tissues during restorative care. | 0.85 | 0.96 | 0.90 | 0.90 | 0.93 |
| LO1 | Dental Implants | Evaluate if a space is suitable for an implant or other tooth replacement. | 0.83 | 0.91 | 0.88 | 0.88 | 0.90 |
| LO2 | | Plan placement/restoration steps considering health, bite, and hygiene. | 0.83 | 0.92 | 0.89 | 0.89 | 0.91 |
| LO3 | | Recognize common problems (looseness, wear, tissue issues) and plan follow-up. | 0.85 | 0.92 | 0.89 | 0.89 | 0.90 |
| LO1 | Endodontics | Recognize signs of tooth nerve/pulp problems and pain patterns. | 0.85 | 0.94 | 0.90 | 0.90 | 0.92 |
| LO2 | | Choose and interpret basic tests; make a simple diagnosis. | 0.84 | 0.91 | 0.87 | 0.86 | 0.88 |
| LO3 | | Plan cleaning/sealing or refer when complex; explain after-care and outcomes. | 0.83 | 0.92 | 0.88 | 0.88 | 0.90 |
| LO1 | Periodontitis | Explain how gum disease develops and the main risk factors. | 0.87 | 0.94 | 0.90 | 0.90 | 0.92 |
| LO2 | | Identify signs from exam and imaging; judge current status and likely course. | 0.83 | 0.90 | 0.87 | 0.87 | 0.88 |
| LO3 | | Plan prevention, treatment steps, and maintenance visits; explain home care. | 0.84 | 0.91 | 0.87 | 0.86 | 0.88 |

**Table 2.** NLI label distribution within cosine similarity bins.

| Cosine bin | Entailment (%) | Neutral (%) | Contradiction (%) |
|---|---|---|---|
| 0.8–0.9 | 47.9 | 41.2 | 10.9 |
| 0.9–1.0 | 77.1 | 16.8 | 6.1 |

**Table 3.** LDA topic coverage for four dental domains.

| | LDA Topics | | | | | | | | | |
|---|---|---|---|---|---|---|---|---|---|---|
| **Domain Name** | **topic0** | **topic1** | **topic2** | **topic3** | **topic4** | **topic5** | **topic6** | **topic7** | **topic8** | **topic9** |
| Caries | 0.19 | 0.01 | 0.04 | 0.29 | 0.28 | 0.01 | 0.09 | 0.02 | 0.01 | 0.05 |
| Dental Implants | 0.02 | 0.56 | 0.04 | 0.02 | 0.01 | 0.02 | 0.03 | 0.22 | 0.07 | 0.02 |
| Endodontics | 0.03 | 0.01 | 0.20 | 0.04 | 0.02 | 0.04 | 0.42 | 0.02 | 0.20 | 0.03 |
| Periodontitis | 0.01 | 0.03 | 0.06 | 0.03 | 0.03 | 0.31 | 0.03 | 0.01 | 0.02 | 0.46 |

**Table 4.** Expert validation (n = 49): retrieval by cosine similarity and evidence checking by NLI, compared with expert alignment labels.

| Method | Decision rule | n | TP | FP | FN | TN | Precision | Recall | F1 | Accuracy |
|---|---|---|---|---|---|---|---|---|---|---|
| BERT-only | cosine ≥ 0.90 | 49 | 19 | 12 | 3 | 15 | 0.61 | 0.86 | 0.72 | 0.69 |
| NLI-only | entail ≥ 0.50 | 49 | 17 | 13 | 5 | 14 | 0.57 | 0.77 | 0.65 | 0.63 |
| BERT + NLI | cosine ≥ 0.90 AND entail ≥ 0.30 | 49 | 17 | 8 | 5 | 19 | 0.68 | 0.77 | 0.72 | 0.73 |